\documentclass[%
aip,
cp,  
amsmath,amssymb,
reprint,%
]{revtex4-2}

\usepackage{graphicx,subfigure}
\usepackage{dcolumn}
\usepackage{bm}
\usepackage{multirow}    

\usepackage[utf8]{inputenc}
\usepackage[T1]{fontenc}
\usepackage{mathptmx} 

\newcommand\tab[1][1cm]{\hspace*{#1}}
\graphicspath{ {./images/} }

\begin{document}
	
	\title{A Comparative Analysis of the Ensemble Methods for Drug Design }
	
	\author{ Rifkat Davronov} 
	\email[Corresponding author: ]{rifqat.davronov@mathinst.uz}	
	
	\author{Fatima Adilova}
	\email{fatadilova@mathinst.uz}
	\affiliation{
		Institute of Mathematics Academy of Sciences, Tashkent, Uzbekistan
	}

	\begin{abstract}
		Quantitative structure-activity relationship (QSAR) is a computer modeling technique for identifying relationships between the structural properties of chemical compounds and biological activity. QSAR modeling is necessary for drug discovery, but it has many limitations. Ensemble-based machine learning approaches have been used to overcome limitations and generate reliable predictions. Ensemble learning creates a set of diverse models and combines them.
		In our comparative analysis, each ensemble algorithm was paired with each of the basic algorithms, but the basic algorithms were also investigated separately. In this configuration, 57 algorithms were developed and compared on 4 different datasets.		
		Thus, a technique for complex ensemble method is proposed that builds diversified models and integrates them. The proposed individual models did not show impressive results as a unified model, but it was considered the most important predictor when combined. We assessed whether ensembles always give better results than individual algorithms. The Python code written to get experimental results in this article has been uploaded to Github (https://github.com/rifqat/Comparative-Analysis).
		
	\end{abstract}
	
	\maketitle
	
	\section{\label{sec:level1}Introduction}
		Quantitative structure-activity relationship (QSAR) is a computer or mathematical modeling method for identifying the relationship between biological activity and the structural properties of chemical compounds. The underlying principle is that variations in structural properties cause different biological activities \textbf{[1]}. Structural properties refer to physicochemical properties, and biological activity corresponds to various pharmacokinetic properties, such as absorption, distribution, metabolism, excretion and toxicity.
		
		QSAR simulations help rank the hit line of a large number of chemicals in terms of their desired biological activity and greatly reduces the number of candidates for testing. QSAR modeling has become a common process in pharmacology, but with all the progress of QSAR, even after the well-known article by a group of co-authors \textbf{[2]} , there are many limitations \textbf{[3, 4]}. 
		
		For example: the data may include more than hundreds of thousands of compounds, or, on the contrary, a very small sample; each compound can be represented by multiple descriptors; some features are highly correlated; it is assumed that the dataset contains some errors as relationships are estimated through in-situ experiments. Due to these and other limitations for predicting a QSAR based model, it is difficult to achieve a reliable prediction result.
		
		In forecasting based on QSAR, machine learning approaches were applied: linear regression models [5] and Bayesian neural networks [6-8] were used. The random forest (RF) [9,10] deserves special mention - it is the most frequently used algorithm with a high level of predictability, simplicity, and reliability. RF is a kind of ensemble method based on sets of decision trees that can prevent overfitting. RF is considered the gold standard in this area, so new QSAR forecasting methods are often compared in performance with this algorithm.
		
		The well-known Merck Kaggle competition in 2012 drew people's attention to neural networks. The winning team used multitasking neural networks (MTNN) [11]. The fundamental learning structure is based on simple feedforward neural networks; it avoids overfitting by studying multiple biological analyzes at the same time. The team achieved results that consistently outperformed the random forest algorithm. Despite achieving high performance with a multitasking neural network, this team ended up using an ensemble combining different methods.
		RF, and many of the algorithms in the famous Kaggle competition used ensemble learning, a technique that creates a set of training models and combines multiple models to produce final predictions. It has been shown theoretically and empirically that the predictive power of ensemble learning is superior to the predictive power of an individual algorithm even if the last are accurate and diverse [12-15]. Ensemble learning manages the strengths and weaknesses of learning individual algorithms, similar to consensus decision making in critical situations.
		
		Thus, it is popular to use algorithm ensembles by using several algorithms and combining their results. In ensembles, the base algorithms generate partially dependent or independent results on the same or a different part of a dataset, and then results are combined in several ways. The success of an ensemble depends on two main properties: the first is the individual success of the base algorithms of the ensemble, and the second one is the independence of base algorithms' results from each other (low error, high diversity)[16].
		
		Ensemble methods, including an ensemble of neural networks based on bootstrap sampling in QSAR (data sampling ensemble) [17]; ensemble versus different training methods for drug interactions [18], Bayesian ensemble model with various QSAR instruments (ensemble method) [8], ensemble training based on qualitative and quantitative SAR models [19], hybrid QSAR prediction model with various training methods [20 ], ensembles with different boosting methods [21], hybrid feature selection and training in QSAR simulations [22], and ensemble against various chemicals to predict carcinogenicity (representative ensembles) [23] have been widely used in drug-like studies.
		
		In contrast to the work in which the results of a comparative analysis of ensemble algorithms are presented  [24],this study aims at overcoming the difficulties of QSAR modeling by using ensembles. Our experiments focus on regression ensembles because this type of models is simpler and easier to understand for medical chemists.  The performance of ensemble algorithms is investigated with respect to ensemble algorithms themselves, and the base algorithms used within the ensemble algorithms.
		
		The article consists of the following sections: in section 2 we described ensemble and base regression algorithms, dimension reduction process, dataset collection. In section 3 presented the results of simulation running and their discussion. In conclude part (section 4) we showed the previous works in order to detailed the success of our results in comparing with achievements  from different studies.
		
	\section{\label{sec:level1} Materials and Methods}
	
		In this section, the base and ensemble algorithms used in our study are briefly described. For the evaluation of the algorithms, the scikit-learn library was used [25]. Each ensemble algorithm was used with each of the base algorithms. The base algorithms were also used alone. With this configuration (2 ensemble + 1 single) x (19 base) = 57 different algorithms were obtained and used.
		
		We ran our experiments on Windows 10 (Intel(R) Core(TM) i7-9700 CPU © 3.00GHz 3.00 GHz). We used the Scikit-learn library package (version 0.23.2) for conventional machine learning methods.

		\subsection{\label{sec:level1} Ensemble algorithms}
	
			Bagging/bootstrapping (BG): Bagging generates N new equal-sized datasets from the original dataset by selecting samples with a replacement [26]. The base algorithms are trained with the datasets. The independence of the individual results is confirmed in the experiments to some degree. N was chosen as 10 in our experiments. The results of the base algorithms are simply averaged to produce the ensemble result.
	
			Additive regression (AR): This is the adaptation of the AdaBoost algorithm to regression types of problems [27]. At each iteration, the samples having big errors at the previous iteration are considered. The iteration number was chosen as 10 in our study. The ensemble result is the weighted mean of the base algorithms. The weights are inversely proportional to the errors of the base algorithms.
	
		\subsection{\label{sec:level2} Base regression algorithms}
			In our study, 19 regression algorithms were used as base learners in the ensembles. They are as follows:
	
			Lasso: The Lasso is a linear model that estimates sparse coefficients. It is useful in some contexts due to its tendency to prefer solutions with fewer non-zero coefficients, effectively reducing the number of features upon which the given solution is dependent. For this reason Lasso and its variants are fundamental to the field of compressed sensing. Under certain conditions, it can recover the exact set of non-zero coefficients [26].
		
			Ridge: This model solves a regression model where the loss function is the linear least squares function and regularization is given by the l2-norm. Also known as Ridge Regression or Tikhonov regularization. This estimator has built-in support for multi-variate regression[27].
	
			ElasticNet: ElasticNet is a linear regression model trained with both $ \ell$1 and $ \ell$2-norm regularization of the coefficients. This combination allows for learning a sparse model where few of the weights are non-zero like Lasso, while still maintaining the regularization properties of Ridge[28].
	
			Orthogonal Matching Pursuit (OMP): OMP is based on a greedy algorithm that includes at each step the atom most highly correlated with the current residual. It is similar to the simpler matching pursuit (MP) method, but better in that at each iteration, the residual is recomputed using an orthogonal projection on the space of the previously chosen dictionary elements[29].
	
			Bayesian Regression: Bayesian regression techniques can be used to include regularization parameters in the estimation procedure: the regularization parameter is not set in a hard sense but tuned to the data at hand[30].
	
			Automatic Relevance Determination: Fit the weights of a regression model, using an ARD prior. The weights of the regression model are assumed to be in Gaussian distributions. Also estimate the parameters lambda (precisions of the distributions of the weights) and alpha (precision of the distribution of the noise). The estimation is done by an iterative procedures (Evidence Maximization)[31].
	
			Passive Aggressive Algorithms: The passive-aggressive algorithms are a family of algorithms for large-scale learning. They are similar to the Perceptron in that they do not require a learning rate. However, contrary to the Perceptron, they include a regularization parameter C [32].
	
			Theil-Sen estimator: TheilSenRegressor is comparable to the Ordinary Least Squares (OLS) in terms of asymptotic efficiency and as an unbiased estimator. In contrast to OLS, Theil-Sen is a non-parametric method which means it makes no assumption about the underlying distribution of the data. Since Theil-Sen is a median-based estimator, it is more robust against corrupted data aka outliers. In univariate setting, Theil-Sen has a breakdown point of about 29.3\% in case of a simple linear regression which means that it can tolerate arbitrary corrupted data of up to 29.3\% [33].
	
			Huber Regression: This makes sure that the loss function is not heavily influenced by the outliers while not completely ignoring their effect[34].
	
			Kernel ridge regression (KRR): Kernel ridge regression combines ridge regression (linear least squares with l2-norm regularization) with the kernel trick. It thus learns a linear function in the space induced by the respective kernel and the data. For non-linear kernels, this corresponds to a non-linear function in the original space [35].
	
			Support Vector Regression (SVR): The model produced by support vector classification (as described above) depends only on a subset of the training data, because the cost function for building the model does not care about training points that lie beyond the margin. Analogously, the model produced by Support Vector Regression depends only on a subset of the training data, because the cost function ignores samples whose prediction is close to their target [36].
	
			Decision Tree Regressor (DTR): The goal is to create a model that predicts the value of a target variable by learning simple decision rules inferred from the data features [37].
	
			Random Forest Regressora: Random forest is a meta estimator that fits a number of classifying decision trees on various sub-samples of the dataset and uses averaging to improve the predictive accuracy and control over-fitting[38]. 
	
			Extra-Trees Regressor: This class implements a meta estimator that fits a number of randomized decision trees  various sub-samples of the dataset and uses averaging to improve the predictive accuracy and control over-fitting[39].
	
			Nearest Neighbors Regression: Neighbors-based regression can be used in cases where the data labels are continuous rather than discrete variables. The label assigned to a query point is computed based on the mean of the labels of its nearest neighbors [40].
	
			Multi-layer Perceptron regressor(MLPR): MLPRegressor trains iteratively since at each time step the partial derivatives of the loss function with respect to the model parameters are computed to update the parameters. It can also have a regularization term added to the loss function that shrinks model parameters to prevent overfitting. This implementation works with data represented as dense and sparse numpy arrays of floating point values [41].
	
		\subsection{\label{sec:level1} Dimension reduction process}
			Drug design datasets generally have a very large number of features. In our study, the original datasets and their dimensionally reduced versions are used. By doing so, the effects of the feature selection process on the accuracies of the algorithms are investigated. The accuracies over the original and dimensionally reduced datasets are compared. The Random Forest Importances method is used for feature selection [42]. 		
		
		\subsection{\label{sec:level1} Dataset collection}
		Our drug data collection consists of 4 drug datasets obtained from several studies. The datasets are shown in Table 1. The datasets with 2075 descriptors were formed using the Dragon [43]. The molecules and outputs were obtained from the original studies.
		
		\begin{table}
		\caption{\label{tab:table1}					
		The 4 original datasets used in study }
		\begin{ruledtabular}
		\begin{tabular}{cclllc}
		Dataset ID & Dataset name & Number of samples & Original number of descriptors   & Number of selected features & Reference \\
		\hline
		1 &	polymer\_133 & \multicolumn{1}{c} {133} & \multicolumn{1}{c}{836} & \multicolumn{1}{c}{10} & [44] \\
		2 & alkaloid\_53 & \multicolumn{1}{c} {53} & \multicolumn{1}{c}{2221} & \multicolumn{1}{c}{10} & [45] \\
		3 & alkaloid\_103 & \multicolumn{1}{c}{103} & \multicolumn{1}{c}{355} & \multicolumn{1}{c}{10} & [46] \\
		4 & Polymer\_150 & \multicolumn{1}{c}{150} & \multicolumn{1}{c}{474} & \multicolumn{1}{c}{10} & [47] \\
		\end{tabular}
		\end{ruledtabular}
		\end{table}
	
	\section{Results and Discussion}
		Nineteen base regressors were used together with each ensemble algorithm on 4 regression-type drug design problems. Before simulation we have posed the same questions as in the article [24], since these are the most important characteristics of ensemble methods:
		\begin{itemize}
			\item Do the algorithm ensembles generate more successful results than a single algorithm?
			\item What is the most successful ensemble algorithm?
			\item What is the base algorithm-ensemble pair with the best results?
			\item Which algorithm performs well with the ensembles?
			\item What is the most successful single algorithm?
			\item How are the algorithms and datasets grouped according to their performances?
			\item How does the dimension reduction process affect the results?
			
		\end{itemize}
	
		To answer these questions, 57 algorithms ((2 ensemble + 1 single) x (19 base algorithms)  = 57) were employed on the 4 drug design datasets described in Table 1 and their dimensionally reduced versions. A cross validation was used and the RMSE results were averaged.
		
		The RMSE is defined as:
		\begin{center}
			
			$RMSE_{a lg10.name}=\sqrt{\frac{1}{N}\displaystyle\sum_{i-1}^{N} y_{a lg10.name}^i-y_{actual}^i }$
			
		\end{center}
		where $y_{a lg10.name}^i$ is the prediction of alg.name for the ith test sample, $y_{actual}^i$ is the actual output value of the ith test sample, and N is the number of test samples.

		Our base and ensemble algorithms have some hyperparameters to optimize. We used the default hyperparameters.
	
  		In the cross-validation methodology, the dataset is randomly divided after shuffling into 2 halves. One half is used in the training and the other is used in the testing. This validation is repeated 5 times. In the results of this validation, 5 estimates of testing the RMSE were obtained for each algorithm and each dataset. In some experiments, very high RMSE results were obtained, especially with the simple linear regression algorithm disturbing the overall averages. Because of this, the performance comparisons of the algorithms were done with the algorithms’ success ranking instead of the averaged RMSEs. In each experiment, the averaged cross-validation RMSEs were sorted in ascending order. The algorithm with the lowest RMSE got the 1st ranking. The worst got the 57th ranking. These success rankings are given in Tables 2 and 3. In Table 2. the results with the original datasets are shown. In Table 3, the results with the dimensionally reduced datasets are shown. The 4 datasets are ordered along the columns of the tables. The algorithms are ordered along the rows of the tables. The average success rate and standard deviation of each algorithm are shown in the last 2 columns.
   
    	In Tables 4 and 5, the summaries of Tables 2 and 3 are given, respectively. Each cell is the averaged success ranking of the experiments with the base algorithm in the cell’s row and the ensemble algorithm in the cell’s column. The average success rankings of the single algorithms used are given in the ‘Single’ column. In the Avg. column, the averaged success rankings of the experiments with respect to the base algorithms are given. In the ‘Avg.’ row, the averaged success rankings of the experiments with respect to the ensemble algorithms are given.
	
		When Tables 2, 3,4,5 are investigated, the following conclusions are reached. For the experiments with the original datasets (Tables 2 and 4):
		\textendash The best ranking performance (5.75) is obtained with the Extra Trees Regressor algorithm.\\
		\textendash The best performed ensemble algorithms are additive regression (AR).\\
		\textendash The best performed base algorithm is Support Vector Machine.\\
		\textendash All of the ensemble algorithms generally increased the performance of each base algorithm. The exceptions are Bayesian Ridge, Support Vector Machine and K Neighbors Regressor.\\
		\textendash The Decision Tree and Support Vector Machine base algorithms had their best performances with BG. The Decision Tree and Automatic Relevance Determination algorithms with AR, achieved their best performances.\\
		\begin{table}
			\caption{\label{tab:table2}	
			 The success ranking of 57 algorithms on 4 original  datasets (see.Table 1) (best to worst, 0 to 56).}
			\begin{ruledtabular}
				\begin{tabular}{lllllll}
				\textbf{No.}& \textbf{Algorithm} & \multicolumn{5}{c}{\textbf{Dataset’s ID}} \\
				\hline
				&  &\textbf{1} & \textbf{2} & \textbf{3} & \textbf{4} & \textbf{Avg.} \\
				\hline
				\textbf{0} & Lasso Regression  & 25 & 2 & 25 & 44 & 24 \\
				
				\textbf{1} & BG-Lasso Regression  & 18 & 6 & 18 & 43 & 21,25 \\
				
				\textbf{2} & AR-Lasso Regression  & 31 & 4 & 31 & 40 & 26,5 \\
				
				\textbf{3} & Ridge Regression	& 52 & 55 &	52 & 27 & 46,5 \\
				
				\textbf{4} & BG-Ridge Regression &	47 & 39 & 47 &	28 & 40,25  \\
				
				\textbf{5} & AR-Ridge Regression &	43 & 29 & 43 &	32 & 36,75 \\
				
				\textbf{6} & Elastic Net &	28 & 1 & 28 & 37 & 23,5 \\
				
				\textbf{7} & BG-Elastic Net &	21 & 3 & 21 & 38 & 20,75 \\
				
				\textbf{8} & AR-Elastic Net &	29 & 0 & 29 & 36 &	23,5 \\
				
				\textbf{9} & Lasso Least Angle Regression &	34 & 44 & 34 &	45 & 39,25 \\
				
				\textbf{10} & BG-Lasso Least Angle Regression &	32 & 43 & 32 & 46 &	38,25 \\
				
				\textbf{11} & AR-Lasso Least Angle Regression &	36 & 42 & 36 & 47 &	40,25 \\
				
				\textbf{12} & Orthogonal Matching Pursuit &	48 & 48 & 48 & 	0  & 36 \\
						
				\textbf{13} &BG-Orthogonal Matching Pursuit	&30&	24&	30&	6&	22,5 \\
				
				\textbf{14} & AR-Orthogonal Matching Pursuit &	24&	19&	24&	1&	17 \\
				
				\textbf{15} & Bayesian Ridge &	26 & 5 & 26 & 29 &	21,5 \\
				
				\textbf{16} & BG-Bayesian Ridge &	49 & 35 & 49 &	23 &	39 \\
				
				\textbf{17} & AR-Bayesian Ridge &	41 & 32 &	41 &	30 &	36 \\
				
				\textbf{18} & Automatic Relevance Determination & 45 &	50 & 45 &	4 &	36 \\
				
				\textbf{19} & BG-Automatic Relevance Determination & 38 &	17 & 38 & 3 &	24 \\
				
				\textbf{20} & AR-Automatic Relevance Determination & 20 & 13 &	20 & 5	& 14,5 \\
				
				\textbf{21} & Passive Aggressive Regressor &	37&	40&	37&	51&	41,25 \\
				
				\textbf{22} & BG-Passive Aggressive Regressor &	35&	30&	35&	52&	38 \\
				
				\textbf{23} & AR-Passive Aggressive Regressor &	40&	14&	40&	53&	36,75 \\
				
				\textbf{24} & TheilSen Regressor &	55&	56&	55&	2&	42 \\
				
				\textbf{25} & BG-TheilSen Regressor &	53&	53&	53&	12&	42,75  \\
				
				\textbf{26} & AR-TheilSen Regressor &	44&	34&	44&	24&	36,5 \\
				
				\textbf{27} & Huber Regressor &	23&	52&	23&	48&	36,5 \\
				
				\textbf{28} & BG-Huber Regressor	& 27&	49&	27&	49&	38 \\
				
				\textbf{29} & AR-Huber Regressor &	33&	46&	33&	50&	40,5 \\
				
				\textbf{30} & Kernel Ridge &	51&	54&	51&	25&	45,25\\
				
				\textbf{31} & BG-Kernel Ridge &	46&	47&	46&	26&	41,25  \\
				
				\textbf{32} & AR-Kernel Ridge &	42&	31&	42&	31&	36,5 \\
				
				\textbf{33} & Support Vector Machine	 & 4&	16&	4&	41&	16,25 \\
				
				\textbf{34} & BG-Support Vector Machine &	1&	23&	1&	39&	16 \\
				
				\textbf{35} & AR-Support Vector Machine &	14&	26&	14&	42&	24 \\
				
				\textbf{36} & K Neighbors Regressor &	19&	41&	19&	33&	28 \\
				
				\textbf{37} & BG-K Neighbors Regressor &	16&	37&	16&	34&	25,75 \\
				
				\textbf{38} & AR-K Neighbors Regressor &	22&	38&	22&	35&	29,25 \\
				
				\textbf{39} & Decision Tree &	39&	8&	39&	19&	26,25 \\
				
				\textbf{40} & BG-Decision Tree &	15&	10&	15&	20&	15  \\
				
				\textbf{41} & AR-Decision Tree &	0&	18&	0&	15&	\textbf{8,25} \\
				
				\textbf{42} & Random Forest	& 12&	12&	12&	18&	13,5 \\
				
				\textbf{43} & BG-Random Forest &	8&	15&	8	&21&	13 \\
				
				\textbf{44}&AR-Random Forest&	13&	28&	13	&11&	16,25\\
					
					\textbf{45}&Extra Trees Regressor&	3&	7&	3&	10&	\textbf{5,75}\\
					
					\textbf{46}&BG-Extra Trees Regressor&	6&	9&	6&	16&	9,25\\
					
					\textbf{47}&AR-Extra Trees Regressor&	11&	20&	11&	14&	14\\
					
					\textbf{48}&AdaBoost Regressor&	9&	25&	9&	17&	15\\
					
					\textbf{49}&BG-AdaBoost Regressor&	2&	21&	2&	22&	11,75\\
					
					\textbf{50}&AR-AdaBoost Regressor&	5&	33&	5&	13&	14\\
					
					\textbf{51}	&Gradient Boosting Regressor&	17&	22&	17&	8&	16\\
					
					\textbf{52}&BG-Gradient Boosting Regressor&	10&	11&	10&	9&	10\\
					
					\textbf{53}&AR-Gradient Boosting Regressor&	7	&27	&7&	7&	12\\
					
					\textbf{54}&Multi Level Perceptron&	56&	51&	56&	56	&54,75\\
					
					\textbf{55}&BG-Multi Level Perceptron&	50&	36&	50&	55&	47,75\\
					
					\textbf{56}&AR-Multi Level Perceptron&	54&	45&	54&	54&	51,75\\
					
				\end{tabular}
			\end{ruledtabular}
		\end{table}
			
    	\begin{table}
	    	\caption{\label{tab:table3}	
	    		The success ranking of 57 algorithms on the 4 dimensionally reduced  datasets (best to worst, 0 to 56).}
	    	\begin{ruledtabular}
	    		\begin{tabular}{lllllll}
    			\textbf{No.}& \textbf{Algorithm} & \multicolumn{5}{c}{\textbf{Dataset’s ID}} \\
    			\hline
    			&  &\textbf{1} & \textbf{2} & \textbf{3} & \textbf{4} & \textbf{Avg.} \\
    			\hline
    			\textbf{0} & Lasso Regression  & 41 & 36 & 41 & 46 & 41 \\
    			
    			\textbf{1} & BG-Lasso Regression  & 38 &32 & 38 &48 & 39 \\
    			
    			\textbf{2} & AR-Lasso Regression  & 37 & 31 & 37 & 45 & 37,5 \\
    			
    			\textbf{3} & Ridge Regression	& 9 & 7 &	9 & 26 & 12,75 \\
    			
    			\textbf{4} & BG-Ridge Regression &12 & 3 & 12 &	23 & 12,5  \\
    			
    			\textbf{5} & AR-Ridge Regression &	19 & 15 & 19 &	17 & 17,5 \\
    			
    			\textbf{6} & Elastic Net &	33& 34 & 33 & 39 & 34,75 \\
    			
    			\textbf{7} & BG-Elastic Net &	31 & 29 & 31 & 40 & 32,75 \\
    			
    			\textbf{8} & AR-Elastic Net &	34 & 28 & 34 & 38 &	33,5 \\
    			
    			\textbf{9} & Lasso Least Angle Regression &	42 & 35 & 42 &	47 & 41,5 \\
    			
    			\textbf{10} & BG-Lasso Least Angle Regression &	39 & 33 & 39 & 49 &	40 \\
    			
    			\textbf{11} & AR-Lasso Least Angle Regression &	45 & 30 & 45 & 50 &	42,5 \\
    			
    			\textbf{12} & Orthogonal Matching Pursuit &	25 & 26 & 25 & 	37  & 28,25 \\
    				
    			\textbf{13} & BG-Orthogonal Matching Pursuit	&44&	12&	44&	36&	34 \\
    			
    			\textbf{14} & AR-Orthogonal Matching Pursuit &	43&	5&	43&	35&	31,5 \\
    			
    			\textbf{15} & Bayesian Ridge &	20 & 21 & 20&9 &	17,5 \\
    			
    			\textbf{16} & BG-Bayesian Ridge &	35& 27 & 35 &	18 &	28,75 \\
    			
    			\textbf{17} & AR-Bayesian Ridge &	30 & 53 &	30 &	5 &	29,5 \\
    			
    			\textbf{18} & Automatic Relevance Determination & 28&	45 & 28&	8&27,25 \\
    			
    			\textbf{19} & BG-Automatic Relevance Determination & 47 &	43 & 47 & 11 &37 \\
    			
    			\textbf{20} & AR-Automatic Relevance Determination & 48 & 52 &	48 & 4	&38 \\
    			
    			\textbf{21} & Passive Aggressive Regressor &	55&	56&	55&	53&	54,75 \\
    			
    			\textbf{22} & BG-Passive Aggressive Regressor &	27&	41&	27&	52&	36,75 \\
    			
    			\textbf{23} & AR-Passive Aggressive Regressor &	29&	54&	29&	51&	40,75 \\
    			
    			\textbf{24} & TheilSen Regressor &	52&	37&	52&	6&	36,75 \\
    			
    			\textbf{25} & BG-TheilSen Regressor &	53&	40&	53&	3&	37,25  \\
    			
    			\textbf{26} & AR-TheilSen Regressor &	54&	44&	54&	7&39,75 \\
    			
    			\textbf{27} & Huber Regressor &	40&	55&	40&	25&	40 \\
    			
    			\textbf{28} & BG-Huber Regressor	& 49&	46&	49&	16&	40 \\
    			
    			\textbf{29} & AR-Huber Regressor &	23&	50&	23&	31&	31,75 \\
    			
    			\textbf{30} & Kernel Ridge &	11&	49&	11&	27&	24,5\\
    			
    			\textbf{31} & BG-Kernel Ridge &	17&	48&	17&	22&	26  \\
    			
    			\textbf{32} & AR-Kernel Ridge &	16&	51&	16&	29&	28 \\
    			
    			\textbf{33} & Support Vector Machine	 & 15&8&15&	43&	20,25 \\    			    			
				    			    			
    			\textbf{34} & BG-Support Vector Machine &	14&	4&	14&	41&	18,25 \\
    			
    			\textbf{35} & AR-Support Vector Machine &	7&	10&	7&	42&	16,5 \\
    			
    			\textbf{36} & K Neighbors Regressor &	36&	13&	36&	33&	29,5 \\
    			
    			\textbf{37} & BG-K Neighbors Regressor &	32&	11&	32&	32&	26,75 \\
    			
    			\textbf{38} & AR-K Neighbors Regressor &	50&	23&	50&	44&	41,75 \\
    			
    			\textbf{39} & Decision Tree &	51&	38&	51&	21&	40,25 \\
    			
    			\textbf{40} & BG-Decision Tree &	46&	24&	46&	24&	35  \\
    			
    			\textbf{41} & AR-Decision Tree &	22&	17&	22&	12&	18,25 \\
    			
    			\textbf{42} & Random Forest	& 24&	18&	24&	10&	19 \\
    			
    			\textbf{43} & BG-Random Forest &	26&	6&	26	&15&	18,25 \\
    			
    			\textbf{44} & AR-Random Forest &	8&	9&	8&	20&	11,25 \\
    			
    			\textbf{45} & Extra Trees Regressor &	0&	2&	0&	0&	\textbf{0,5} \\
    			
    			\textbf{46} & BG-Extra Trees Regressor &	2 &	0&2&	1&	1,25 \\
    			
    			\textbf{47} & AR-Extra Trees Regressor &	1&	1& 1&	2&	1,25 \\
    			
    			\textbf{48} & AdaBoost Regressor &	13&	22&	13&	34&	20,5 \\
    			
    			\textbf{49} & BG-AdaBoost Regressor &	10&	14&	10&	28&	15,5 \\
    			
    			\textbf{50} & AR-AdaBoost Regressor &	4&	16&	4&	30&	13,5 \\
    			
    			\textbf{51} & Gradient Boosting Regressor &	3&	25&	3&	19&	12,5 \\
    			
    			\textbf{52} & BG-Gradient Boosting Regressor &	18&	20&	18&	13&	17,25 \\
    			
    			\textbf{53} & AR-Gradient Boosting Regressor	& 6&	19&	6&	14&	11,25 \\
    			
    			\textbf{54} & Multi Level Perceptron &	56&	47&	56&	56&	53,75 \\
    			
    			\textbf{55} & BG-Multi Level Perceptron &	21&	42&	21&	55&	34,75  \\
    			
    			\textbf{56} & AR-Multi Level Perceptron &	5&	39&	5&	54&	25,75 \\
    			
    			\end{tabular}
	    	\end{ruledtabular}
	    \end{table}

		\begin{table}
			\caption{\label{tab:table4}	
				The averaged success rankings of the algorithms on the original datasets (best to worst, 0 to 56).}
			\begin{ruledtabular}
				\begin{tabular}{lllll}
				\multicolumn{1}{c} {\textbf{Algorithm}}&\multicolumn{1}{c} {\textbf{BG}} & \multicolumn{1}{c} {\textbf{AG}}&\multicolumn{1}{c} {\textbf{Single}}&\multicolumn{1}{c} {\textbf{Avg.}} \\
				\hline
				Lasso Regression&	21,25&	26,5&	24&	23,92\\
				
				Ridge Regression&	40,25&	36,75&	46,5&	41,17\\
				
				Elastic Net	&20,75&	23,5&	23,5&	22,58\\
				
				Lasso Least Angle Regression&	38,25&	40,25&	39,25&	39,25\\
				
				Orthogonal Matching Pursuit&	22,5&	17	&36	&25,17\\
				
				Bayesian Ridge&	39&	36&	21,5&	32,17\\
				
				Automatic Relevance Determination&	24&	14,5&	36&	24,83\\
				
				Passive Aggressive Regressor&	38	&36,75&	41,25&	38,67\\
				
				TheilSen Regressor	&42,75&	36,5&	42&	40,42\\
				
				Huber Regressor&	38	&40,5&	36,5& 38,33\\
				
				Kernel Ridge&	41,25&	36,5&	45,25&	41,00\\
				
				Support Vector Machine&	16&	24&	16,25&	18,75\\
				
				K Neighbors Regressor&	25,75&	29,25&	28&	27,67\\
				
				Decision Tree&	15&	8,25&	26,25&	16,50\\
				
				Random Forest&	13&	16,25&	13,5&	14,25\\
				
				Extra Trees Regressor&	9,25&	14&	5,75&	9,67\\
				
				AdaBoost Regressor	&11,75&	14&	15&	13,58\\
				
				Gradient Boosting Regressor&	10&	12&	16&	12,67\\
				
				Multi Level Perceptron&	47,75&	51,75&	54,75&	51,42\\
				
				Avg.&	27,08	&27,07&	29,86& \\

				\end{tabular}
			\end{ruledtabular}
		\end{table}

		\begin{table}
			\caption{\label{tab:table5}	
				The averaged success rankings of the algorithms on the dimensionally reduced datasets (best to worst, 0 to 56).}
			\begin{ruledtabular}
				\begin{tabular}{lllll}
				\multicolumn{1}{c} {\textbf{Algorithm}}&\multicolumn{1}{c} {\textbf{BG}} & \multicolumn{1}{c} {\textbf{AG}}&\multicolumn{1}{c} {\textbf{Single}}&\multicolumn{1}{c} {\textbf{Avg.}} \\
				\hline
				Lasso Regression&	39&	37,5&	41&	39,17\\
				
				Ridge Regression&	12,5&	17,5&	12,75&	14,25\\
				
				Elastic Net	&32,75&	33,5&	34,75&	33,67\\
				
				Lasso Least Angle Regression&	40&	42,5&	41,5&	41,33\\
				
				Orthogonal Matching Pursuit&	34&	31,5&	28,25&	31,25\\
				Bayesian Ridge&	28,75&	29,5&	17,5&	25,25\\
				Automatic Relevance Determination&	37&	38&	27,25&	34,08\\
				Passive Aggressive Regressor&	36,75&	40,75&	54,75&	44,08\\
				TheilSen Regressor&	37,25&	39,75&	36,75&	37,92\\
				Huber Regressor&	40&	31,75&	40&	37,25\\
				Kernel Ridge&	26&	28&	24,5&	26,17\\
				Support Vector Machine&	18,25&	16,5&	20,25&	18,33\\
				K Neighbors Regressor&	26,75&	41,75&	29,5&	32,67\\
				Decision Tree&	35&	18,25&	40,25&	31,17\\
				Random Forest&	18,25&	11,25&	19&	16,17\\
				Extra Trees Regressor&	1,25&	1,25&	0,5&	1,00\\
				AdaBoost Regressor&	15,5&	13,5&	20,5&	16,50\\
				Gradient Boosting Regressor&	17,25&	11,25&	12,5&	13,67\\
				Multi Level Perceptron&	34,75&	25,75&	53,75&	38,08\\
				Avg.&	27,95&	26,83&	29,22\\

				\end{tabular}
			\end{ruledtabular}
		\end{table}
		For the experiments with the dimensionally reduced datasets (Tables 3 and 5):\\
		\tab\textendash The best ranking performance (0.5) is obtained with the Extra Trees Regressor algorithm.\\
		\tab\textendash The best performed ensemble algorithms are additive regression (AR) and bagging (BG).\\
		\tab\textendash The best performed base algorithm is Ridge Regression.\\
		\tab\textendash All of the ensemble algorithms generally increased the performance of each base algorithm. The exceptions \tab are Orthogonal Matching Pursuit, Bayesian Ridge, Automatic Relevance Determination, TheilSen Regressor \tab and Kernel Ridge.\\
		\tab\textendash The Ridge Regression and Support Vector Machine base algorithms had their best performances with BG. \tab The Ridge Regression, Support Vector Machine and Decision Tree algorithms with AR, achieved their best \tab performances.
		
		The average successes of the algorithms were investigated above. Next, the best performing algorithm will be investigated over each individual dataset. In Table 6, the dataset name, and the error and the name of the best performing algorithm are shown for the original and dimensionally reduced datasets.
		
		\begin{table}
			\caption{\label{tab:table6}	
				The averaged success rankings of the algorithms on the dimensionally reduced datasets (best to worst, 0 to 56).}
			\begin{ruledtabular}
				\begin{tabular}{lllll}
				\multirow{2}{*} &\multicolumn{2}{c} {With all of the features} & \multicolumn{2}{c} {With the selected features} \\
				\hline
                Dataset name & Best performing algorithm & RMSE & Best performing algorithm & RMSE \\
				\hline
				polymer\_133&	AR-Automatic Relevance Determination&	0,01&	BG-Gradient Boosting Regressor&	0,01\\
				
				alkaloid\_53&	AR-Elastic Net&	0,29&	BG-Extra Trees Regressor&	0,31\\
				
				alkaloid\_103&	AR-Decision Tree&	0,57&	Extra Trees Regressor&	0,65\\
				
				Polymer\_150&	Orthogonal Matching Pursuit&	0,00&	Orthogonal Matching Pursuit&	0,02\\				
				
			\end{tabular}
		\end{ruledtabular}
		\end{table}
	
		When Table 6 is investigated, the following conclusions are reached:\\
		\tab\textendash The best performing algorithms are generally ensemble algorithms. This is in agreement with the average \tab success of the algorithms.\\
		\tab\textendash experiments with dimensional reduced data sets do not have better results than the original data sets, except \tab for 1 data set (polymer\_133).
		
		When the algorithms are clustered, the algorithms are represented by points having 4 (the number of datasets) features (dimensions). When the datasets are clustered, the datasets are represented by points having 57 (the number of algorithms) features (dimensions).
		
		According to Figure 1, the following conclusions are reached: \\
		\tab \textendash In both figures, the ensemble-algorithm pairs are generally clustered with their base single algorithms. \\
		\tab \textendash The feature selection process does not affect the similarities of the algorithms dramatically.
		
		According to Figure 2, the following conclusions are reached: \\
		\tab \textendash On the left side of Figure 2, there is no obvious pattern between the clusters and the number of features/samples. \\
		\tab \textendash To the right of Figure 2, the polymers and alkaloids are clustered separately. 
		\newline
		\newline
		\newline
		\newline
		\newline
		\newline
		\newline
		\newline
		\newline
		\newline
		\newline
		\newline
		\newline
		\begin{figure}
			\centering     
			\includegraphics[width=168mm,height=215mm]{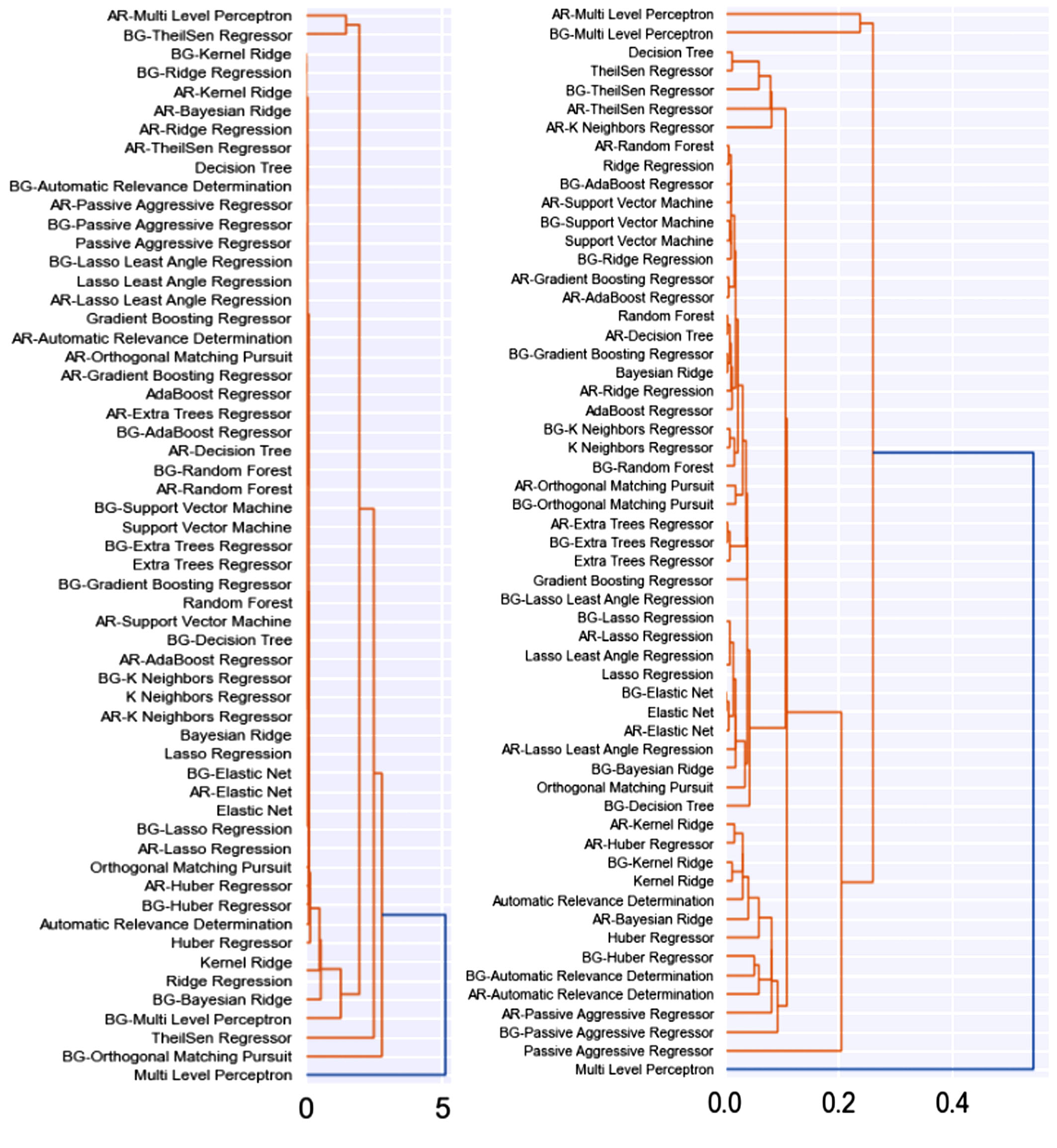}			
			\caption{The hierarchical clusters of the algorithms according to their RMSE values on the original (left) and dimensionally reduced (right) 4 datasets.}
		\end{figure}
		
		\begin{figure}
			\centering     
			\includegraphics[width=168mm,height=150mm]{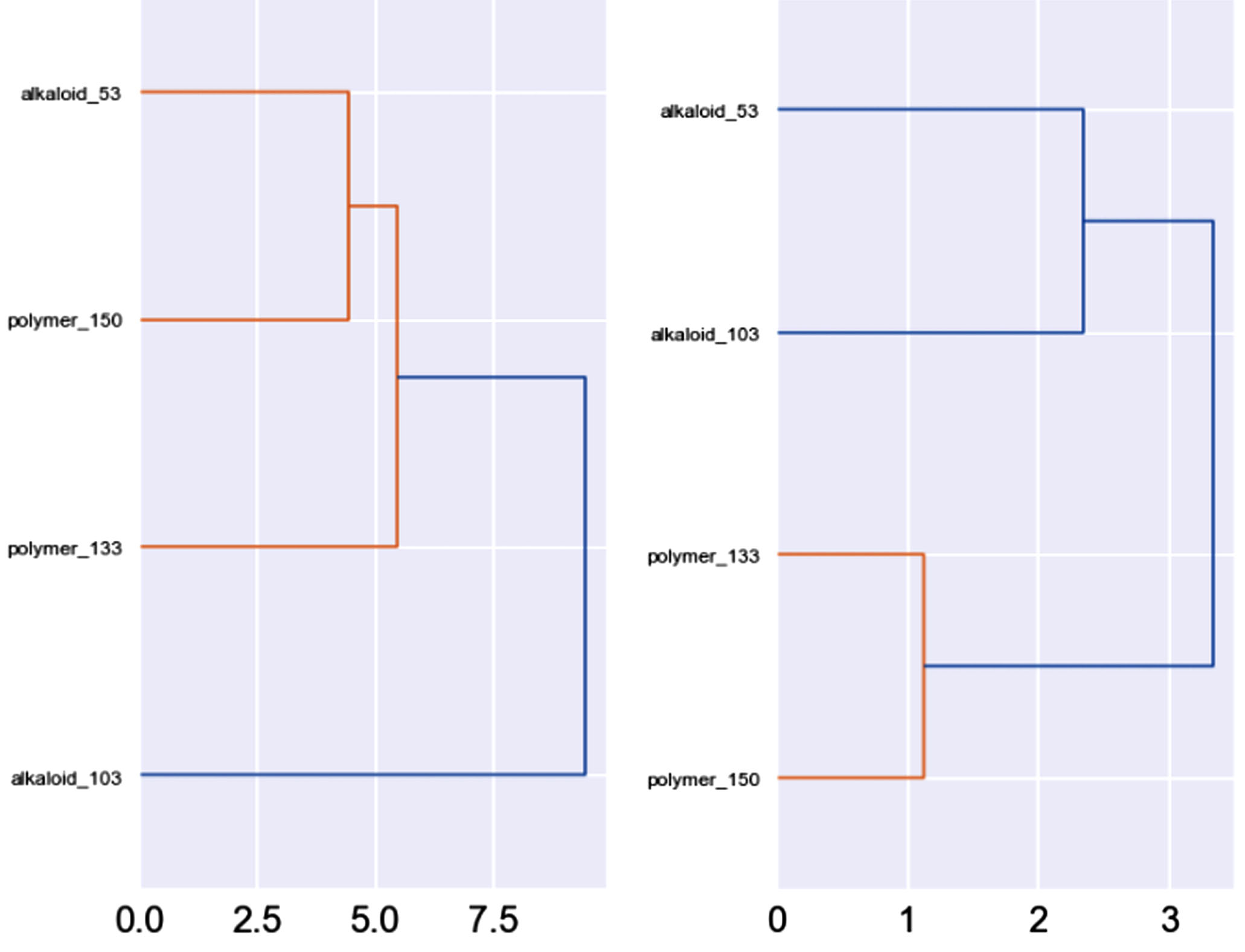}
			\caption{The hierarchical clusters of the original (left) and dimensionally reduced (right) 4 datasets according to their RMSE values obtained with 57 algorithms. In the figures, the dataset names, the number of features, and the samples are given.}
		\end{figure}
	\section{Previous works}
		The selected previous studies in this area for both classification and regression are shown comparatively in Table 7.

		According to Table 7, together with our experiments, the following conclusions are reached:\\
		\tab \textendash 	The number of drug design / chemical data sets used in our experiments is greater than in previous studies, \tab except for [24].\\
		\tab \textendash 	The number of base machine learning methods used in our experiments is greater than in previous studies.\\
		\tab \textendash 	The superior success of ensemble algorithms over single algorithms is confirmed.\\

		\begin{table}
			\caption{\label{tab:table7}	Previous works.}
			\begin{ruledtabular}
				\begin{tabular}{llll}
					
				\multicolumn{1}{c} {\textbf{Reference}}&\multicolumn{1}{c} {\textbf{Compared methods in the study}} & \multicolumn{1}{c} {\textbf{Datasets}}&\multicolumn{1}{c} {\textbf{Results}} \\
				\hline
				[48]&	ctree, rtree, cforest, rforest, gbm, fnn, &	1 regression-type &	RandomForest showed \\
				& earth, glmnet, ridge, lm, pcr, plsr, rsm, & (chemical data). & good results.\\
				&   rvm, ksvm, ksvmfp, nnet, nneth2o & & \\
				\hline

				[24] &(4 ensemble + 1 single) * (7 base) = 35 &	15 regression-type 	&Ensemble methods showed \\
				     &                                        & (chemical data).    & good results.\\
				\hline
				[49] &	AdaBoostM1+Bagging (Ada\_Bag),  
				
				&	& Bagging (Ada\_Bag)   \\
				&AdaBoostM1+Jrip (Ada\_Jrip),& & and Random Forest \\
				&AdaBoostM1+J48 (Ada\_J48),& & (Ada\_RF) algorithms \\
				&AdaBoostM1+PART (Ada\_PART),& & showed good  \\
				&AdaBoostM1+RandomForest (Ada\_RF),&	MDDR database. & results. \\
				&and & &\\
				&AdaBoostM1+REPTree (Ada\_RT). & & \\
				\hline
				[50]& 	EnsemDT, EnsemKRR and 	& 2 classification datasets  &	EnsemDT and EnsemKRR  than \\
				& other single methods & (chemical data) &showed better results\\
				& & & other single methods.\\
				\hline
				
			\end{tabular}
			\end{ruledtabular}
		\end{table}
	
	\section{Conclusion}
		In machine learning, committee algorithms (ensembles), especially those with classification applications, are highly popular because they have better performances than single algorithms.
	
		In this study, the comparative performances of algorithm ensembles with drug design datasets in regression applications were investigated. A drug design dataset collection with 4 regression-type datasets was used for this purpose. We obtained the performances of the single algorithms and the algorithm ensembles on those datasets. The combinations of 19 base algorithms and 2 ensemble algorithms were investigated.

\begin{acknowledgments}
The authors would like to express their sincere gratitude to the Ministry of innovative development of the Republic of Uzbekistan (Grant FA-tech-3018-4). The authors are is also thankful to several colleagues in both laboratories, who were engaged in many scientific discussions that helped the authors formulate the best current practices modeling discussed herein.
\end{acknowledgments}

\nocite{*}
\bibliography{aipsamp_ver3}

\end{document}